\documentclass{article}
\usepackage{ijcai22}

\usepackage{times}
\usepackage{soul}
\usepackage{url}
\usepackage[hidelinks]{hyperref}
\usepackage[utf8]{inputenc}
\usepackage[small]{caption}
\usepackage{graphicx}
\usepackage{amsmath}
\usepackage{amsthm}
\usepackage{booktabs}
\usepackage{algorithm}
\usepackage{algorithmic}
\usepackage{multirow}
\usepackage{subfigure}
\usepackage{amssymb}
\usepackage{amsfonts}
\usepackage{bm}
\usepackage{appendix}

\usepackage{pifont}  
\newcommand{\cmark}{\ding{51}}  
\newcommand{\xmark}{\ding{55}}  

\title{Contrastive Self-Supervised Learning with Hard Negative Pair Mining}
\author{
Wentao Zhu\and 
Hang Shang\and
Tingxun Lv\and
Chao Liao\and
Sen Yang\and
Ji Liu \\
Kuaishou Technology
}

\begin{document}

\maketitle

\begin{abstract}
Recently, learning from vast unlabeled data, especially self-supervised learning, has been emerging and attracted widespread attention. Self-supervised learning followed by the supervised fine-tuning on a few labeled examples can significantly improve label efficiency and outperform standard supervised training using fully annotated data~\cite{chen2020big}. In this work, we present a novel self-supervised deep learning paradigm based on online hard negative pair mining. Specifically, we design a student-teacher network to generate multi-view of the data for self-supervised learning and integrate hard negative pair mining into the training. Then we derive a new triplet-like loss considering both positive sample pairs and mined hard negative sample pairs. Extensive experiments demonstrate the effectiveness of the proposed method and its components on ILSVRC-2012.  
\end{abstract}

\section{Introduction}
Learning from a large scale unlabeled dataset has long been a hot topic in the computer vision community, because the large amount of high quality labels require laborious and costly annotation for each task and there exists huge amount of unlabeled data from various data servers and sources. Un/self-supervised learning can effectively learn a {\it task-agnostic} representation from vast unlabeled data, and downstream tasks, such as image classification, can be well performed by fine-tuning on a few {\it task-specific} labels. This strategy has become a main-stream pipeline for the transformer based self-supervised learning approaches~\cite{vaswani2017attention}. Recent advanced self-supervised learning achieves promising results and outperforms conventional fully supervised learning method on the image classification~\cite{chen2020big}.

The key effort of general self-supervised learning approaches mainly focuses on pretext task construction~\cite{jing2020self}. The pretext task can be designed to be predictive tasks~\cite{mathieu2016deep}, generative tasks~\cite{bansal2018recycle}, contrastive tasks~\cite{oord2018representation}, or a combination of them. The supervision signal for the pretext task, i.e., {\it pseudo label}, typically is yielded from a pretext construction process which generally involves exhausted multi-view construction to model various variations~\cite{qian2020spatiotemporal}. Through solving the pretext task with specific objective function, the network learns transferable visual features for various downstream tasks.  

The study of conventional self-supervised learning methods mainly involves data related pretext task design~\cite{zhu2021test}. Popular pretext tasks include colorizing gray scale images~\cite{zhang2016colorful}, image inpainting~\cite{pathak2016context}, playing image jigsaw puzzle~\cite{noroozi2016unsupervised}, etc. For the video related self-supervised learning approaches, the data related pretext tasks can be sequence order verification~\cite{misra2016shuffle}, solving sequence sorting~\cite{lee2017unsupervised}, predicting the odd or unrelated element~\cite{fernando2017self}, classifying clip order~\cite{xu2019self}, etc.

Recent tremendous success of self-supervised learning is mainly introduced by advanced learning strategies. The InfoNCE loss is widely adopted for contrastive learning, which maximizes a lower bound of mutual information based on the {\it pseudo label} in the pretext task~\cite{oord2018representation}. SimCLR employs larger batch sizes, more training steps and composition of data augmentations, which matches the performance of a fully supervised ResNet-50 simply by adding one additional linear classifier~\cite{chen2020simple}. \cite{wu2018unsupervised} maintains a large feature memory bank to store training image representation. MoCo builds a large and consistent dictionary through a dynamic queue and a momentum-updated encoder, which outperforms its supervised pretraining counterpart in the detection and segmentation~\cite{he2020momentum}. SimSiam employs a stop-gradient operation in the Siamese architectures to prevent collapsing solutions of self-supervised learning~\cite{chen2020exploring}. SimCLRv2 employs big (deep and wide) networks during pretraining and fine-tuning, and it achieves surprising good performance for semi-supervised learning on ImageNet~\cite{chen2020big}. BYOL trains an online network to predict a target network representation of the same image where the target network is a slow-moving average of the online network~\cite{grill2020bootstrap}.  

In this work, we propose a novel self-supervised learning paradigm by introducing an effective negative image pair mining in the contrastive learning framework. Specifically, we introduce a student-teacher network into the contrastive learning framework to construct multi-view representation of data. To effectively learn from unlabeled data in the contrastive learning, we further construct the negative image pairs by hard negative image pair mining. The overall objective function can be derived as a form of triplet-like loss facilitated by the collected positive and negative image pairs.  

We conduct extensive experiments including linear evaluation, semi-supervised learning, transfer learning, and ablation study to evaluate our method on the ImageNet dataset~\cite{russakovsky2015imagenet}. The proposed method achieves 77.1\% top-1 accuracy using a ResNet-50 encoder for the linear evaluation, which outperforms previous state-of-the-art by 2.8\%. For the semi-supervised learning task, our method with a ResNet-50 encoder obtains the top-1 accuracy of 73.4\%, which outperforms previous best result by 4.6\% using 10\% labels. For the transfer learning with linear evaluation, our method with a ResNet-50 encoder achieves the best accuracy on six out of seven widely used transfer learning datasets, which averagely outperforms previous best results by 2.5\%.
More specifically, our major contributions are summarized as follows.
\begin{itemize}
\item First, we build a student-teacher network to construct multi-view representations in the contrastive learning framework. The gradient of the student sub-network is blocked to ease the training difficulty and stabilize the training of self-supervised learning. 
\item Second, we collect hard negative image pairs on-the-fly and add the hard negative image pairs into the training of contrastive self-supervised learning. 
\item Third, extensive experiments demonstrate the adversarial contrastive self-supervised learning outperforms previous state-of-the-art self-supervised learning approaches for linear evaluation, semi-supervised learning and transfer learning on the ImageNet dataset.
\end{itemize}

\section{Related Work}
The mainstream unsupervised/self-supervised learning literature generally involves two aspects: data/feature related pretext tasks and loss functions~\cite{he2020momentum}. The data/feature related pretext tasks typically can be specially constructed by the multi-view data/feature generation process~\cite{jing2020self}. Through solving the pretext task, the deep network of self-supervised learning is expected to learn a good representation for the down-stream tasks. Loss objective functions can often improve the performance of self-supervised learning significantly. The adversarial contrastive learning focuses on the novel loss function based on advanced student-teacher network design. Next we discuss related study with respect to these aspects.


Contrastive loss measures the similarity of image pairs in the feature space~\cite{hadsell2006dimensionality}. In contrastive learning framework, the target can be defined and generated on-the-fly during training~\cite{hadsell2006dimensionality}. Recent significant success of self-supervised learning has witnessed the widespread adoption of contrastive learning~\cite{henaff2020data}. \cite{zhuang2019local} train an embedding function to maximize a metric of local aggregation, causing similar data instances to move together in the embedding space, while allowing dissimilar instances to separate. Contrastive multi-view learning trains deep network by maximizing mutual information between different views of the same scene~\cite{tian2019contrastive}. 

The student-teacher network can be used to generate multi-view representations of unlabeled data. Temporal ensembling maintains an exponential moving average (EMA) prediction as the {\it pseudo label} for the self-supervised training~\cite{laine2016temporal}. Instead of averaging label predictions, mean-teacher uses EMA to update model weights~\cite{tarvainen2017mean}. MoCo further uses a momentum to update the encoder for the new keys on-the-fly, and maintains a queue of keys in the contrastive learning framework~\cite{he2020momentum}. BYOL maintains a student-teacher network to yield multi-view of samples in the training~\cite{grill2020bootstrap}. Without negative sample pairs in the training,  BYOL achieves surprisingly good performance. Momentum teacher performs two independent momentum updates for teacher’s weight and teacher’s batch normalization statistics to maintain a stable training process~\cite{li2021momentum}.  

\section{Method}

\begin{figure}[t]
\begin{center}
\includegraphics[width=0.5\textwidth]{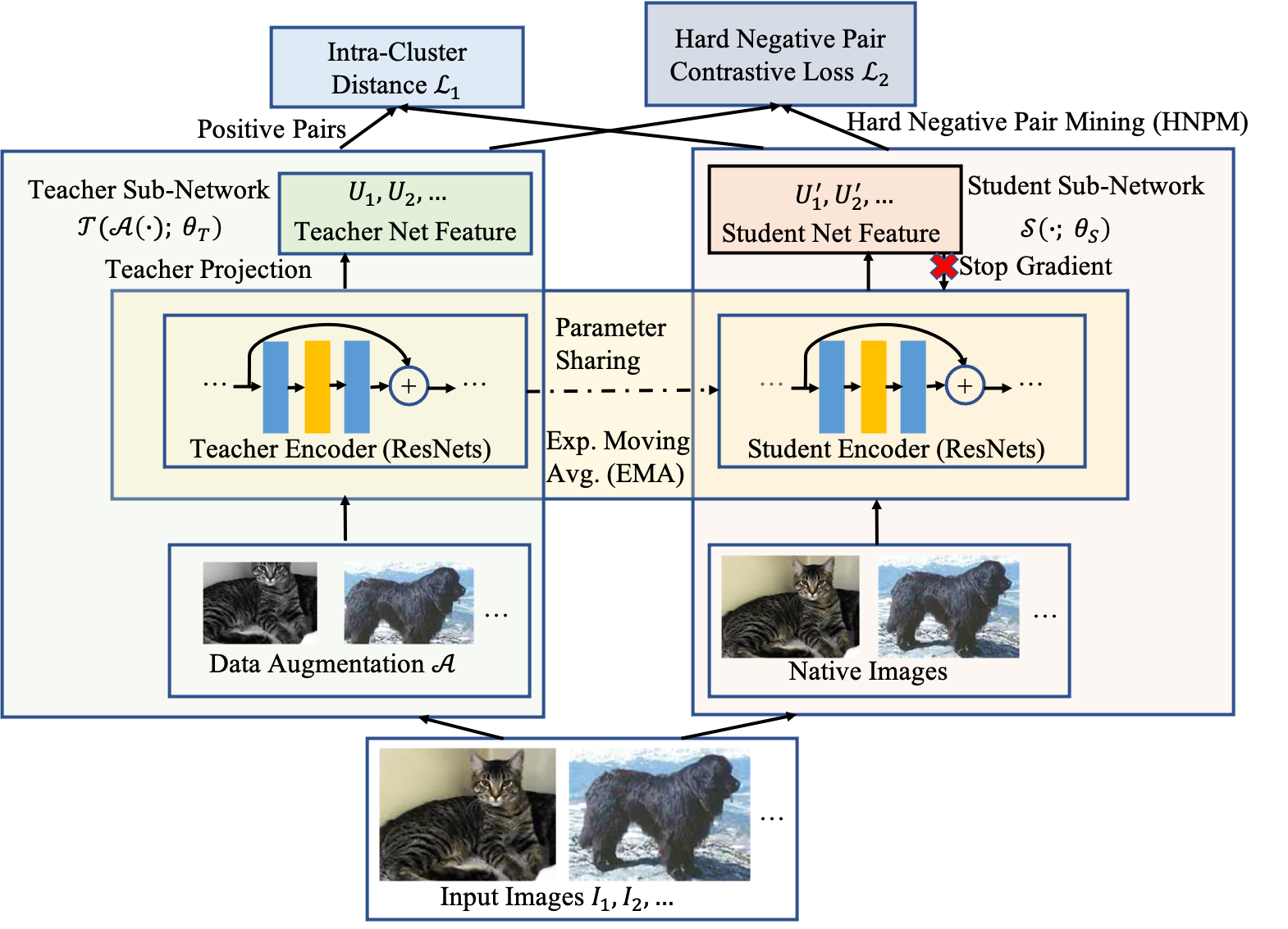}
\end{center}
\caption{The architecture of contrastive self-supervised learning with hard negative pair mining.}
\label{fig:example}
\end{figure}

We employ a student-teacher network to construct two representational views of the sample, as illustrated in Fig.~\ref{fig:example}. At the top of the student and teacher sub-networks, we construct both the positive sample pairs and negative sample pairs. Specifically, we consider the representations of the same sample from student and teacher sub-networks as the positive pair, and we only retain the most similar pair of two different samples to construct the negative pair, i.e., hard negative pair. We block the gradient update of the student sub-network and employ the exponential moving average (EMA) to update its parameters to stabilize the self-supervised training.  

\subsection{Student-Teacher Network}\label{sec:stunet}
\textbf{Problem definition:} Un/self-supervised learning tries to learn a good representation from a large scale unlabeled dataset $\mathcal{D} = \{I_1, I_2, \cdots, I_N\}$, where each $I$ represents an image. For an image sampled from the dataset $I_i \sim \mathcal{D}$, we can obtain two representational views of $I_i$ by constructing one student sub-network $\mathcal{S}\left( \cdot; {\theta}_S \right)$ and one teacher sub-network $\mathcal{T}(\cdot; {\theta}_T)$~\cite{shin2020semi}. To let the network learn various invariant, we employ advanced data augmentation $\mathcal{A}$, including color jittering, horizontal flipping, Gaussian blurring and random cropping, in the data generation process for the teacher sub-network. We then obtain two views of representations for image $I_i$ as
\begin{equation}
    {U}_{i} = \mathcal{T}\left(\mathcal{A}\left(I_i\right); \theta_T\right), \quad
    U_i^{\prime} = \mathcal{S}(I_i; \theta_S),
\end{equation}
where $U_i$ is the representation from the teacher sub-network and ${U}^{\prime}_{i}$ is the representation from the student sub-network.

The self-supervised learning tries to build pretext tasks from these unlabeled data. The generated representation views $U_i$ and ${U}^{\prime}_{i}$ from the student and teacher sub-networks can be considered as a positive pair, which belong to the same cluster. The adversarial contrastive self-supervised learning tries to yield a compact representation for images of the same cluster by minimizing their normalized $L_2$ distance in the representational space. The intra-cluster distance can be defined
\begin{equation}
\begin{aligned}
\mathcal{L}_1 = \mathbb{E}_{I_i \sim \mathcal{D}} {\left[ \left( \frac{U_i}{\| U_i \|_{\infty}} - \frac{U^{\prime}_i}{\| U^{\prime}_i \|_{\infty}} \right)^2 \right]},
\end{aligned}\label{eq:pos}
\end{equation}
where images are randomly sampled from the dataset $I_i \sim \mathcal{D}$, $\|\cdot \|_{\infty}$ is the infinity norm, i.e., the maximum of the absolute value of elements in the vector.

\subsection{Hard Negative Pair Mining (HNPM)}\label{sec:hard}
It is not efficient to train a self-supervised network by solely using positive pairs of samples. Current self-supervised learning uses large batch size~\cite{chen2020simple}, memory bank~\cite{wu2018unsupervised} or large dynamic dictionary~\cite{he2020momentum} to achieve promising results.  Adding negative image pairs can significantly improve the training efficiency of a self-supervised learning model.

We heuristically construct negative pairs in the self-supervised learning framework by mining hard negative pairs of images. For two different image $I_i$ and image $I_j$, we measure the dissimilarity of the two images by the normalized $L_2$ distance in the representation space
\begin{equation}
    \begin{aligned}  
    & U_j = \mathcal{T}\left(\mathcal{A}\left(I_j\right); \theta_T\right), \\
    & \text{DisSim}(U_i^{\prime}, U_j) = \left( \frac{U_i^{\prime}}{\| U_i^{\prime} \|_{\infty}} - \frac{U_j}{\| U_j \|_{\infty}} \right)^2.
    \end{aligned}\label{eq:dissim}
\end{equation}

There exists large numbers of negative pairs of samples. Hard samples have been widely proved to improve the performance of a deep learning model~\cite{ren2015faster,lin2017focal}. In the self-supervised learning framework, we define the hard negative pairs to be image pairs of small dissimilarity according to Equation~\ref{eq:dissim}. We try to maximize the normalized $L_2$ distance or dissimilarity of negative image pairs. The contrastive loss for negative pairs can be derived  
\begin{equation}
    \begin{aligned}
    \mathcal{L}_2 = - \mathbb{E}_{I_i \sim \mathcal{D}} [ \log \big(\sum_{I_j \in \Tilde{\mathcal{B}}_i} (\text{DisSim}(U_i^{\prime}, U_j))  \big) ], 
    \end{aligned}\label{eq:neg}
\end{equation}
where images are randomly sampled from the dataset $I_i \sim \mathcal{D}$, $\Tilde{\mathcal{B}}_i$ is the hard negative sample set of the current batch $\mathcal{B}_i$ for image $I_i$. The hard negative sample set $\Tilde{\mathcal{B}}_i$ can be constructed  
\begin{equation}
    \begin{aligned}
    \Tilde{\mathcal{B}}_i = \{I_j | I_j \in \mathcal{B}_i, I_j \neq I_i,  \text{DisSim}(U_i^{\prime}, U_j) \leq 1 \}.
    \end{aligned}\label{eq:hard_neg}
\end{equation}
We construct hard negative pairs on-the-fly in training, which can be used to efficiently train the self-supervised network.

\subsection{Network Update}\label{sec:netupdate}
To stabilize the training and avoid a collapsing solution in the self-supervised learning~\cite{chen2020exploring}, we block the gradient for the student sub-network $\mathcal{S}(\cdot; \theta_S)$. We employ the exponential moving average (EMA) to update the parameters $\theta_S$ in the student sub-network~\cite{tarvainen2017mean}
\begin{equation}
\begin{aligned}
\theta_S \leftarrow \tau \theta_S + (1 - \tau) * \theta_T, 
\end{aligned}\label{eq:stu_update}
\end{equation}
where $\tau$ is a smoothing coefficient to tune the update strength of the student sub-network.

In the back-propagation, we only use the gradient to update the parameters of the teacher sub-network. The overall loss function can be derived as
\begin{equation}
\begin{aligned}
\mathcal{L}(\theta_T) = \alpha_1 \mathcal{L}_1 + \alpha_2 \mathcal{L}_2,
\end{aligned}\label{eq:loss}
\end{equation}
where $ 0 < \alpha_1 < 1$ and $0 < \alpha_2 < 1$ are the fixed coefficients to tune the trade-off between the intra-cluster loss and inter-cluster loss. During the back-propagation, we employ the gradient clipping to stabilize the training.  

\subsection{Connection with InfoNCE and Stability}\label{sec:info}
In our method, we employ hard negative pair mining (HNPM) to add negative image pairs in the training, and use a normalized $L_2$ distance in the loss function. We will demonstrate that minimizing the loss of our method is equivalent with minimizing the InfoNCE loss~\cite{oord2018representation}. To simplify the analysis, we temporarily remove the hard negative pair mining mechanism in our method in the derivation of connection with InfoNCE.

The InfoNCE loss~\cite{oord2018representation} can be written as
\begin{equation}
    \begin{aligned}
\mathcal{L}_{NCE} &= -\mathbb{E}_{I_i \sim \mathcal{D}}[\log \frac{f_k (U_i, U_i^{\prime})}{\sum_{I_j \in \mathcal{D}} f_k (U_j, U_i^{\prime})}].
\end{aligned}
\end{equation}
where $U_i$, $U_i^{\prime}$ are calculated from the teacher sub-network and the student sub-network, $f_k(\cdot, \cdot)$ models the mutual information between the encoded representations in the InfoNCE and can use similarity loss as a surrogate loss to approximate the mutual information.

We define the similarity loss as the reciprocal of normalized $L_2$ distance of the encoded representations. The InfoNCE loss can then be defined as
\begin{equation}
    \begin{aligned}
\mathcal{L}_{NCE} &\triangleq \mathbb{E}_{I_i \sim \mathcal{D}}[\log \frac{\text{DisSim} (U_i, U_i^{\prime})}{\sum_{I_j \in \mathcal{D}} \text{DisSim} (U_j, U_i^{\prime})}] \\
&= \mathbb{E}_{I_i \sim \mathcal{D}}[\log  (\frac{U_i}{\| U_i \|_{\infty}} - \frac{U_i^{\prime}}{\| U_i^{\prime} \|_{\infty}})^2] \\  & \quad - \mathbb{E}_{I_i \sim \mathcal{D}} [ \log \big( \sum_{I_j \in \mathcal{D}} (\frac{U_j}{\| U_j \|_{\infty}} - \frac{U_i^{\prime}}{\| U_i^{\prime} \|_{\infty}})^2 \big)].
\end{aligned}\label{eq:nce_sim}
\end{equation}
The second part of derived loss in equation~\ref{eq:nce_sim} is the same with our negative pair loss in the equation~\ref{eq:neg} if we temporarily neglect our hard negative sample pair mining for each batch. Minimizing the first part of equation~\ref{eq:nce_sim} is equivalent with minimizing $\mathbb{E}_{I_i \sim \mathcal{D}}[  (\frac{U_i}{\| U_i \|_{\infty}} - \frac{U_i^{\prime}}{\| U_i^{\prime} \|_{\infty}})^2] $, which is the positive pair loss in the equation~\ref{eq:pos}. From the above derivation, we conclude, with the proper relaxation and assumption, minimizing the our loss is equivalent with minimizing the InfoNCE loss.

Next we demonstrate that the hard negative pair mining (HNPM) leads to stable training. Without the trade-off factors $\alpha_1$ and $\alpha_2$, the loss can be written as
\begin{equation}
    \begin{aligned}
    \mathcal{L} &= \mathbb{E}_{I_i \sim \mathcal{D}}[  (\frac{U_i}{\| U_i \|_{\infty}} - \frac{U_i^{\prime}}{\| U_i^{\prime} \|_{\infty}})^2] \\
    &\quad - \mathbb{E}_{I_i \sim \mathcal{D}} [ \log \big( \sum_{I_j \in \Tilde{\mathcal{B}}_i} (\frac{U_j}{\| U_j \|_{\infty}} - \frac{U_i^{\prime}}{\| U_i^{\prime} \|_{\infty}})^2 \big)].
    \end{aligned}
\end{equation}
Without loss of generality, we remove the normalization constraint and denote $\frac{U_i}{\| U_i \|_{\infty}}$ as $U_i$. 
\begin{equation}
    \begin{aligned}
    \mathcal{L} &= \mathbb{E}_{I_i \sim \mathcal{D}} \big[  ( U_i - U_i^{\prime} )^2 - \log \big( \sum_{I_j \in \Tilde{\mathcal{B}_i}} ( U_j - U_i^{\prime})^2 \big) \big].
    \end{aligned}
\end{equation}

The hard negative pair mining (HNPM) always explores negative pairs with $L_2$ distance smaller than 1, which guarantees $( U_j - U_i^{\prime})^2$ is bounded to be smaller than 1. We use $M$ to denote the upper bound of negative pair loss. 
\begin{equation}
    \begin{aligned}
    | \mathcal{L} | &\leq \mathbb{E}_{I_i \sim \mathcal{D}} \big[  ( U_i - U_i^{\prime} )^2 \big] + M.
    \end{aligned}\label{eq:bound}
\end{equation}

Next we can further prove equation~\ref{eq:bound} can be optimized stably, and the first part of equation~\ref{eq:bound}, i.e., the loss of positive pairs, can be decreased consecutively by escaping undesirable equilibria. If the model stacks into an undesirable equilibrium solution, the feature representation of teacher sub-network can be denoted as $\mathbb{E}[U_i^{\prime} | U_i]$ from the update rule in equation~\ref{eq:stu_update}. The loss of positive pairs $\mathcal{L}_P$ can be derived as
\begin{equation}
    \begin{aligned}
    \mathcal{L}_P &= \mathbb{E}_{I_i \sim \mathcal{D}} \big[  ( U_i - U_i^{\prime} )^2 \big] \\
    & = \mathbb{E}_{I_i \sim \mathcal{D}} \big[  ( \mathbb{E}[U_i^{\prime} | U_i] - U_i^{\prime} )^2 \big] = \mathbb{E}_{I_i \sim \mathcal{D}} [Var(U_i^{\prime} | U_i)].
    \end{aligned}
\end{equation}
Let $Z$ denote an additional variability induced by stochasticities in the training dynamics. We always have a solution leading to a lower loss during the training, which escapes the current equilibrium, because
\begin{equation}
    \begin{aligned}
    Var(U_i^{\prime} | U_i, Z) \leq  Var(U_i^{\prime} | U_i).
    \end{aligned}
\end{equation}
From the above derivation, the learning is stable with the benefit of hard negative pair mining and student sub-network updating rule.

\subsection{Implementation Details}\label{sec:imple}
Because of our advanced learning strategy, we do not use any pretrained model as the backbone in our implementation. To generate multi-view representations, we employ data augmentation to model various variations in different views.

We use residual networks as the student sub-network $\mathcal{S}(\cdot, \theta_{S})$ and teacher sub-network $\mathcal{T}(\cdot, \theta_{T})$. The two coefficients of the loss in equation~\ref{eq:loss}, $\alpha_1$ is set to 0.8 and $\alpha_2$ is set to 0.1. We employ the gradient clipping strategy in the back-propagation where we set the maximum norm of gradient clipping as $1.0$. The Adam optimizer is used to minimize the loss in equation~\ref{eq:loss}. The batch size is 160. The learning rate is set as $0.1$ and we use a cosine annealing schedule for the learning rate with the maximum number of iterations as $100$. The smoothing coefficient $\tau$ in the update of student sub-network in equation~\ref{eq:stu_update} is set as $0.5$.  

We employ data augmentation for teacher sub-network on-the-fly during training. We firstly apply color jittering with brightness of $0.8$, contrast of $0.8$, saturation of $0.8$, and hue of $0.2$ to random 80\% training images in each batch. Then we convert random 20\% images to gray scale, and horizontally flip 50\% images. After that, we smooth random 10\% images with a random Gaussian kernel of size $3\times 3$ and standard deviation of $1.5 \times 1.5$. Finally, we crop each image with random crop size of scale range $[0.8, 1.0]$. We use the mean of $[0.485, 0.456, 0.406]$ and the standard deviation of $[0.229, 0.224, 0.225]$ to normalize RGB channels.  

\section{Experiments}
We conduct experiments to validate the performance of the proposed method on the ILSVRC-2012 dataset.  

\subsection{Linear Evaluation}
\begin{table}[]
    \begin{center}
    \begin{tabular}{c|c|c}
    \hline
        Method & Top-1 & Top-5 \\
        \hline\hline
        CPCv2~\cite{henaff2020data} & 63.8 & 85.3 \\
        CMC~\cite{tian2019contrastive} & 66.2 & 87.0 \\
        SimCLR~\cite{chen2020simple} & 69.3 & 89.0 \\
        MoCov2~\cite{chen2020improved} & 71.1 & N/A \\
        SimCLRv2~\cite{chen2020big} & 71.7 & N/A \\
        InfoMin Aug.~\cite{tian2020makes} & 73.0 & 91.1 \\
        BYOL~\cite{grill2020bootstrap} & 74.3 & 91.6 \\
        Ours & \textbf{77.1} & \textbf{93.7}\\  
        \hline
    \end{tabular}
    \end{center}
    \caption{\label{tab:lineval_res50}The accuracy comparison of self-supervised learning (SSL) approaches with the ResNet-50 encoder based on linear evaluation on the ImageNet dataset. The bold face denotes the best accuracy.}
\end{table}

{\small
\begin{table}[]
    \begin{center}
    \begin{tabular}{c|c|c|c|c}
        \hline
        Method & Dep. & Wid. & Top-1 & Top-5 \\
        \hline\hline
        CMC & 50 & 2$\times$ & 70.6 & 89.7 \\
        SimCLRv2 & 50 & 2$\times$ & 75.6 & N/A \\
        BYOL & 50 & 2$\times$ & 77.4 & 93.6 \\
        Ours & 50 & 2$\times$ & \textbf{79.4} & \textbf{94.5} \\ \hline
        SimCLR & 50 & 4$\times$ & 76.5 & 93.2 \\
        BYOL & 50 & 4$\times$ & 78.6 & 94.2 \\
        Ours & 50 & 4$\times$ & \textbf{80.3} & \textbf{95.1} \\ \hline
        BYOL & 200 & 2$\times$ & 79.6 & 94.8\\
        Ours & 200 & 2$\times$ & \textbf{81.9} & \textbf{96.4} \\
        \hline
    \end{tabular}
    \end{center}
    \caption{\label{tab:lineval}The accuracy (\%) comparison of SSL methods with other ResNet encoders based on linear evaluation.}  
\end{table}}
The linear evaluation can be used to evaluate the accuracy of self-supervised learning (SSL) by freezing the SSL model and training a separate linear classifier after the SSL model~\cite{grill2020bootstrap,kornblith2019better,zhang2016colorful}. We compare our method with previous state-of-the-art approaches with the ResNet-50 encoder and other ResNet encoders on ImageNet in the Table~\ref{tab:lineval_res50} and Table~\ref{tab:lineval}, respectively. The top-1 and top-5 accuracy are listed. With the standard ResNet-50 encoder~\cite{he2016deep}, our method obtains 77.1\% top-1 accuracy and 93.7\% top-5 accuracy, which outperform previous state-of-the-art top-1 and top-5 results by 2.8\% and 2.1\%, respectively. Most surprisingly, our method achieves 0.6\% better accuracy than the accuracy, 76.5\%, of the supervised baseline from SimCLR~\cite{chen2020simple}.

Table~\ref{tab:lineval} reports the accuracy of self-supervised learning methods using deeper and wider ResNet encoders based on linear evaluation. Our method with ResNet-200 (2$\times$) obtains 81.9\% top-1 and 96.4\% top-5 accuracy which increase previous best top-1 and top-5 accuracy by 2.3\% and 1.6\%, respectively. With ResNet-50 (2$\times$) and ResNet-50 (4$\times$) encoders, our method also achieves better accuracy than those of CMC~\cite{tian2019contrastive}, SimCLRv2~\cite{chen2020big} and BYOL~\cite{grill2020bootstrap} with the same encoder.

\subsection{Semi-Supervised Learning}
\begin{table*}[]
    \begin{center}
    \begin{tabular}{c|c|c||c|c}
    \hline
        Method & Top-1 (1\%) &  Top-5 (1\%) &Top-1 (10\%) & Top-5 (10\%) \\
        \hline\hline
        SimCLR~\cite{chen2020simple} & 48.3 & 75.5 & 65.6 & 87.8 \\
        SimCLRv2~\cite{chen2020big} & \textbf{57.9} &  N/A &68.4 & N/A \\
        BYOL~\cite{grill2020bootstrap} & 53.2 &  78.4 &68.8 & 89.0 \\
        Ours & 56.7 &  \textbf{80.2} (1.8$\uparrow$) & \textbf{73.4} (4.6$\uparrow$) &\textbf{92.5} (3.5$\uparrow$)\\
        \hline
    \end{tabular}
    \end{center}
    \caption{The accuracy (\%) comparison of SSL methods with the ResNet-50 encoder based on semi-supervised learning on ImageNet dataset.}
    \label{tab:l1}
\end{table*}

\begin{table*}[]
    \begin{center}
    \begin{tabular}{c|c|c|c|c|c|c||c|c}
    \hline
        Method & Dep. & Wid. & SK & Para. & Top-1 & Top-5 & Top-1 (10\%) & Top-5 (10\%) \\
        \hline\hline
        SimCLR~\cite{chen2020simple} & 50 &2$\times$& \xmark & 94M & 58.5 & 83.0 & 71.7 & 91.2 \\
        BYOL~\cite{grill2020bootstrap} & 50&2$\times$ &\xmark& 94M & 62.2  & 84.1 & 73.5 & 91.7 \\
        Ours & 50&2$\times$ & \xmark&94M & \textbf{65.7} &  \textbf{86.2} &\textbf{78.6} (5.1$\uparrow$) & \textbf{93.2} (1.5$\uparrow$) \\ \hline
        SimCLR~\cite{chen2020simple} & 50&4$\times$& \xmark& 375M & 63.0 & 85.8 & 74.4& 92.6 \\
        BYOL~\cite{grill2020bootstrap} & 50&4$\times$& \xmark& 375M & 69.1  & 87.9 & 75.7& 92.5 \\
        Ours & 50&4$\times$& \xmark& 375M & \textbf{70.3} & \textbf{89.9} & \textbf{78.9} (3.2$\uparrow$) & \textbf{95.5} (2.9$\uparrow$) \\ \hline
        BYOL~\cite{grill2020bootstrap} & 200&2$\times$& \xmark & 250M & 71.2 & 87.9 & 77.7& 92.5 \\
        Ours & 200&2$\times$& \xmark & 250M & \textbf{76.5} & \textbf{90.3} & \textbf{80.7} (3.0$\uparrow$) & \textbf{95.4} (2.9$\uparrow$)\\ \hline
        SimCLRv2 distilled~\cite{chen2020big}{} & 50 &1$\times$& \xmark& N/A & 73.9  & 91.5 & 77.5 & 93.4 \\
        SimCLRv2 distilled~\cite{chen2020big} & 50 &2$\times$&\cmark & N/A & 75.9  & 93.0 & 80.2& 95.0 \\
        SimCLRv2 self-distilled~\cite{chen2020big} & 152 &3$\times$&\cmark & N/A & 76.6  & 93.4 & 80.9& 95.5 \\
        Ours & 152 &3$\times$&\cmark & N/A & \textbf{77.6}  & \textbf{94.2}& \textbf{81.3} & \textbf{95.7}  \\
        \hline
        
    \end{tabular}
    \end{center}
    \caption{The accuracy (\%) comparison of SSL approaches with other ResNet encoders including selective kernel convolution (SK) based on semi-supervised learning on the ImageNet dataset.}
    \label{tab:my_label}
\end{table*} 

Semi-supervised learning can also be used to evaluate the accuracy of self-supervised learning (SSL) by fine-tuning representation with a small subset of the training set~\cite{grill2020bootstrap}. In this experiment, we use the fixed data splits of 1\% and 10\% of the training set in ImageNet, which are the same as~\cite{grill2020bootstrap}. We also use the top-1 and top-5 accuracy as the evaluation metric for the semi-supervised learning. The comparison using the ResNet-50 encoder and deeper and wider ResNet encoders are listed in Table~\ref{tab:l1} and Table~\ref{tab:my_label}, respectively.  Our method achieves 80.2\% top-5 accuracy based on a ResNet-50 encoder which improves previous best result by 1.8\% using only 1\% training labels in the Table~\ref{tab:l1}. Using 10\% training labels, our method achieves 73.4\% and 92.5\% for the top-1 and top-5 accuracy, which improve previous best top-1 and top-5 accuracy by 4.6\% and 3.5\%.

The result with ResNet of various depths, widths, selective kernel convolution~\cite{li2019selective} configurations are listed in Table~\ref{tab:my_label}. Our method achieves the best top-1 and top-5 accuracy for all the experimental configurations. Specifically, based on ResNet-50 (2$\times$) encoder, our method achieves 65.7\% and 78.6\% top-1 accuracy using 1\% training labels and 10\% training labels, which improves previous best top-1 accuracy by 3.5\% and 5.1\%. Based on ResNet-200 (2$\times$), our method obtains 76.5\% and 80.7\% top-1 accuracy using 1\% training labels and 10\% training labels, which improves the accuracy of BYOL~\cite{grill2020bootstrap} by 5.3\% and 3.0\%.

\subsection{Transfer Learning}

\begin{table*}[h!]
    \begin{center}
    \begin{tabular}{c|c|c|c|c|c|c|c}
    \hline
        Method & Food101 & CIFAR-10 & SUN397 & Cars & Pets & VOC 2007 & Flowers \\
        \hline \hline
        BYOL~\cite{grill2020bootstrap} & 75.3 & 91.3 & 60.6 & 67.8 & 90.4 & 82.5 & 96.1 \\
        SimCLR~\cite{chen2020simple} & 68.4 & 90.6 & 58.8 & 50.3 & 83.6 & 80.5 & 91.2 \\
        Supervised-IN~\cite{chen2020simple} & 72.3 & \textbf{93.6} & 61.9 & 66.7 & 91.5 & 82.8 & 94.7 \\
        Ours & \textbf{77.6} & 92.4 & \textbf{65.4} & \textbf{72.3} & \textbf{94.2} & \textbf{87.7} & \textbf{97.0}  \\
        \hline
    \end{tabular}
    \end{center}
    \caption{{\label{tab:transfer_linear}}The transfer learning accuracy (\%) comparison of SSL approaches with ResNet-50 encoder based on linear evaluation on ImageNet.}
\end{table*}

\begin{table*}[h!]
    \begin{center}
    \begin{tabular}{c|c|c|c|c|c|c|c}
    \hline
        Method & Food101 & CIFAR-10 & SUN397 & Cars & Pets & VOC 2007 & Flowers \\
        \hline \hline
        BYOL~\cite{grill2020bootstrap} & 88.5 & 97.8 & 63.7 & 91.6 & 91.7 & \textbf{85.4} & 97.0 \\
        SimCLR~\cite{chen2020simple} & 88.2 & 97.7 & 63.5 & 91.3 & 89.2 & 84.1 & 97.0 \\
        Supervised-IN~\cite{chen2020simple} & 88.3 & 97.5 & \textbf{64.3} & \textbf{92.1} & 92.1 & 85.0 & \textbf{97.6} \\
        Ours & \textbf{89.1} & \textbf{98.0} & 64.1 & \textbf{92.1} & \textbf{92.8} & 85.3 & 97.5 \\
        \hline
    \end{tabular}
    \end{center}
    \caption{{\label{tab:transfer_fine}}The transfer learning accuracy (\%) comparison of SSL approaches with the ResNet-50 encoder based on finetuning on ImageNet.}
\end{table*}

Transfer learning is another widely used task to evaluate the accuracy of self-supervised learning (SSL) methods. Transfer learning can be used to evaluate the generalization ability of the learned SSL model. Practically, both linear evaluation, i.e., only training the last classification layer, and fine-tuning the whole network based on the target dataset can be employed for the evaluation of transfer learning. The comparison of transfer learning with linear evaluation and fine-tuning are listed in the Table~\ref{tab:transfer_linear} and Table~\ref{tab:transfer_fine}.

For the linear evaluation of transfer learning task, our method achieves better accuracy than previous state-of-the-art approaches on six out of seven widely used transfer learning datasets in Table~\ref{tab:transfer_linear}. We provide the accuracy improvement in Table~\ref{tab:transfer_linear}, and our method improves 2.3\%, 3.5\%, 4.5\%, 2.7\%, 4.9\% and 0.9\% on Flood101, SUN397, Cars, Pets, VOC 2007 and Flowers datasets, respectively. On average, the transfer learning accuracy of our method is 2.5\% higher than previous best results based on linear evaluation. For the transfer learning with a fine-tuning task, our method achieves the best accuracy on four out of seven tasks in Table~\ref{tab:transfer_fine}.

\begin{figure}[t]
    \begin{center}
    \includegraphics[width=0.23\textwidth]{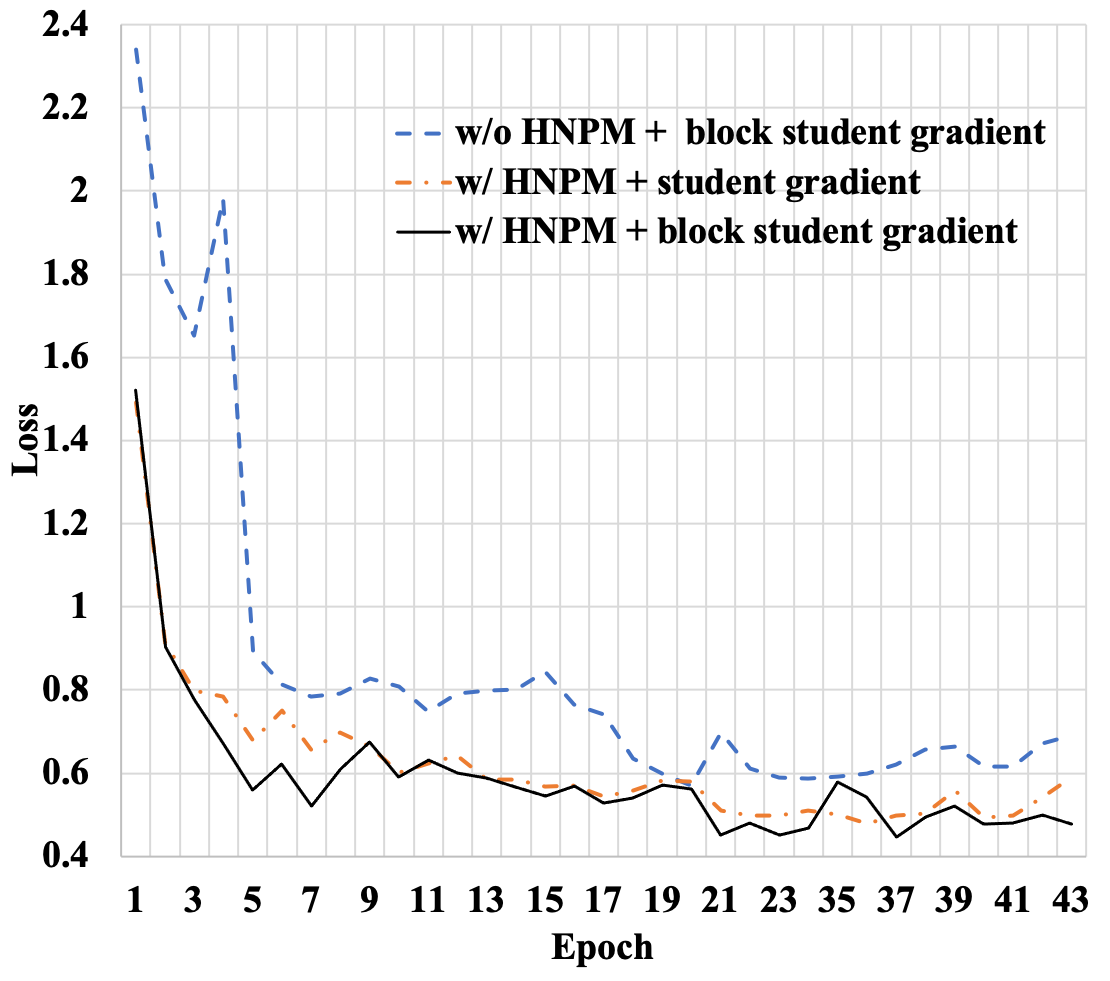}
    \includegraphics[width=0.23\textwidth]{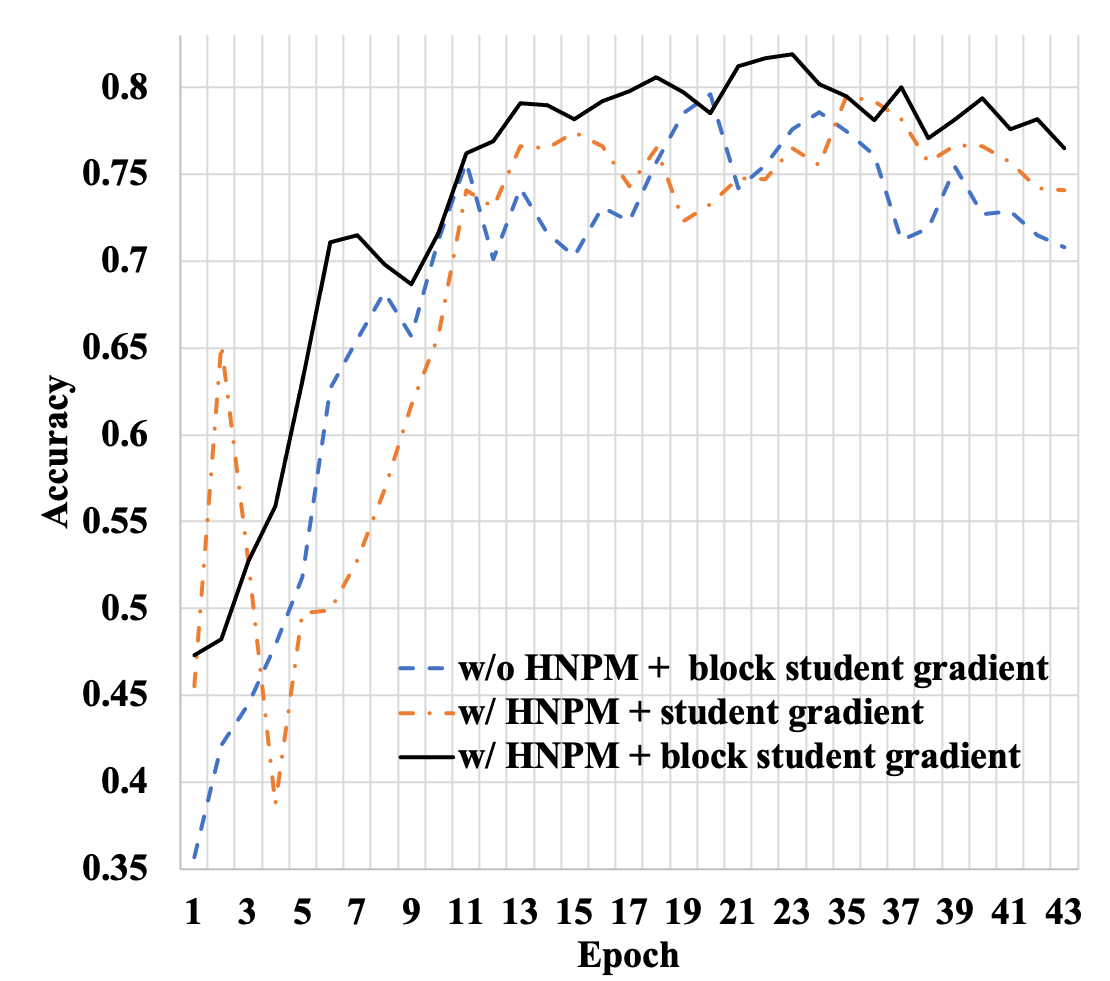}
    \end{center}
    \caption{The loss (left) and accuracy (right) comparison $w.r.t.$ different epochs for ablation study of hard negative pair mining (HNPM) and blocking gradient in the student sub-network based on linear evaluation with the ResNet-200 (2$\times$) encoder on ImageNet.}
    \label{fig:loss_ablation}
\end{figure}


\subsection{Ablation Study}\label{sec:ablation}

\begin{table}[]
    \begin{center}
    \begin{tabular}{c|c|c|c|c}
    \hline
        $\tau$ & 1.0 & 0.999 & 0.5 & 0.0 \\ 
        \hline
        Top-1 (\%) & 24 & 73.4 & \textbf{77.1} & 49.1 \\
        \hline
    \end{tabular}
    \end{center}
    \caption{{\label{tab:ema}}The effect of the smoothing coefficient $\tau$ in the exponential moving average with ResNet-50 encoder based  on linear evaluation.}
\end{table}

\textbf{Coefficient $\tau$ in the update of student sub-network} We investigate the accuracy of our method with linear evaluation based on the ResNet-50 encoder with respective to the smoothing coefficient $\tau$ of the exponential moving average (EMA) in Table~\ref{tab:ema}. The bigger the $\tau$ is, the smaller update the student sub-network performs. When the $\tau$ is 0, it means that we copy the weights of the teacher sub-network to update the student sub-network completely in each step. When the $\tau$ is 1, it means that the student sub-network is never updated. We find that at moving average coefficient value of 0.5, we obtain the best top-1 accuracy, 77.1\%, based on linear evaluation. Neither the moving average coefficient $\tau$ of 0 nor 1 generates good performance.

\textbf{Hard negative pair mining (HNPM)} We conduct ablation study on hard negative pair mining (HNPM) based on liner evaluation task using the ResNet-200 (2$\times$) encoder on the ImageNet dataset. Training with all negative pairs, i.e., without HNPM, is denoted as ``w/o HNPM + block student gradient'', and our method is trained with HNPM, which is denoted as ``w/ HNPM + block student gradient''. The loss and accuracy comparison $w.r.t.$ the training epochs for the two methods are in Fig.~\ref{fig:loss_ablation}. With hard negative pair mining, the training of our method is much more stable and it achieves lower loss and higher accuracy than that without hard negative pair mining.


\textbf{Blocking gradient of student sub-network} We also conduct ablation study on blocking gradient of student sub-network in Fig.~\ref{fig:loss_ablation}. Training without blocking the gradient of student sub-network is denoted as ``w/ HNPM + student gradient''. Our method achieves lower loss and higher accuracy than that with gradient updating of student sub-network. 

\section{Conclusion}
In this work, we introduce a self-supervised learning framework in a student-teacher network with contrastive loss. To increase the training efficiency, we add the hard negative image pairs into the contrastive self-supervised learning paradigm. To stabilize the training and avoid a collapsing solution, we block the gradient of student sub-network and update the parameters of the student sub-network using exponential moving average. We also conduct ablation study to validate the effectiveness of each component. Extensive experiments demonstrate that our method achieves better performance than previous state-of-the-art approaches based on linear evaluation, semi-supervised learning and transfer learning on the ImageNet dataset.  

\bibliographystyle{named}
\bibliography{maskfusion}

\clearpage

\appendix


\end{document}